\documentclass[letterpaper, 10 pt, conference]{ieeeconf}  

\IEEEoverridecommandlockouts                              

\usepackage{graphics} 
\usepackage[caption=false,font=footnotesize]{subfig}
\usepackage{epsfig} 
\usepackage{mathptmx} 
\usepackage{times} 
\usepackage{amsmath} 
\usepackage{amssymb}  
\usepackage{cite}
\usepackage[T1]{fontenc}

\newif\ifarxiv
\arxivtrue 

\ifarxiv
\usepackage{tikz}
\fi

\def\ie{\textit{i.e.,\ }}
\def\eg{\textit{e.g.,\ }}

\title{\LARGE \bf
Simulation of Surgical Suturing Using Position-Based Dynamics and the Material Point Method for Robot Reinforcement Learning
}

\author{Tleukhan Mussin$^{1,*}$, Yafei Ou$^{1,*}$, and Mahdi Tavakoli$^{1,2}$
\thanks{This research was supported by the Canada Foundation for Innovation (CFI), the Natural Sciences and Engineering Research Council (NSERC) of Canada, the Canadian Institutes of Health Research (CIHR), and Alberta Innovates. \textit{(Corresponding author: Yafei Ou.)}}
\thanks{$^{1}$Tleukhan Mussin, Yafei Ou, and Mahdi Tavakoli are with the Department of Electrical and Computer Engineering, University of Alberta, Edmonton, Alberta, Canada (e-mail: {\tt\small yafei.ou@ualberta.ca}).}
\thanks{$^{2}$Mahdi Tavakoli is also with the Department of Biomedical Engineering, University of Alberta, Edmonton, Alberta, Canada.}%
\thanks{$^{*}$Tleukhan Mussin and Yafei Ou contributed equally to this work.}
}

\begin{document}

\bstctlcite{IEEEexample:BSTcontrol}

\maketitle
\thispagestyle{empty}
\pagestyle{empty}

\ifarxiv
\begin{tikzpicture}[remember picture, overlay]
\node [align=left, xshift=10cm, yshift=-0.8cm] at (current page.north west) 
{
\begin{minipage}{19cm} 
\footnotesize
\textcopyright 2026 IEEE.  Personal use of this material is permitted.  Permission from IEEE must be obtained for all other uses, in any current or future media, including reprinting/republishing this material for advertising or promotional purposes, creating new collective works, for resale or redistribution to servers or lists, or reuse of any copyrighted component of this work in other works.
\end{minipage}
};
\end{tikzpicture}
\fi

\begin{abstract}

Recent advances in robotics research have created a strong demand for high-performance simulators. Surgical robotics simulation faces unique challenges due to the need to model diverse objects, such as rigid instruments, soft tissue, and fluids. While many studies simulate sutures or soft tissue independently, only a few have considered the complete soft-tissue suturing scenario, including the contact between sutures and deformable tissue during suture insertion.
Building on previous work, this paper presents a novel suturing simulation environment using sutures modelled by position-based dynamics (PBD) and soft bodies modelled by the material point method (MPM) while considering two-way contact with frictional and drag forces.
We introduce a contact coupling method between the PBD suture and the MPM soft tissue, enabling visually plausible suture-tissue interactions.
The simulator is optimized for GPU execution with parallel scenes using multiple CUDA streams, and we present a Reinforcement Learning (RL) environment for autonomous suturing sub-tasks, including needle insertion, driving, and extraction. Using ML-Agents, RL agents trained in the simulator show stable learning and achieve 80\% and 68\% success rates in needle insertion and extraction, respectively, under the strictest distance threshold.

\end{abstract}

\section{INTRODUCTION}
\label{sec:introduction}

Recent advances in robotics research have created a strong demand for high-performance, high-fidelity simulators, which serve both as playgrounds for testing robotics control and automation algorithms and as sources for synthesizing virtual training data and experiences.
However, despite the availability of many robotics simulation platforms, surgical robotics simulation faces unique challenges due to the need to model diverse objects, such as rigid instruments, soft tissue, and fluids.

The modelling of rope and rod-like objects in surgical simulators, particularly sutures, has been explored in existing studies.
Surgical simulators and computer graphics literature typically represent such objects using rigid-body chains, mass-spring systems, position-based dynamics (PBD), or Cosserat rod formulations to achieve real-time performance \cite{guebert2009SuturingSimulation,maciel2011SurgicalModelviewcontroller,yu2020RealtimeSuturing,munawar2022OpenSimulationa,kim2025SurgicalRobotics,westergaard2026RealtimeHapticbased}.
For example, the Asynchronous Multi Body Framework (AMBF) implements a sequential impulse (SI) solver for simulating a chain of rigid bodies for representing sutures~\cite{munawar2022OpenSimulationa}.
In \cite{westergaard2026RealtimeHapticbased}, extended PBD (XPBD) is used for both suture and soft tissue simulation in a VR-based surgical simulator.
Some studies have also explored using the finite-element method (FEM), although such implementations are expected to be more computationally expensive and less numerically stable \cite{guebert2009SuturingSimulation}.
In a more recent study \cite{kim2025SurgicalRobotics}, Isaac Sim is used for building surgical simulation scenes, where three different methods for modelling suture threads are compared, including using rigid-body chain, FEM deformable body, and PBD particle body.
Recent studies have explored training autonomous suturing agents in simulators \cite{wu2024SurgicAIHierarchical,yu2024OrbitSurgicalOpenSimulationa}.

\begin{figure}[t]
    \centering
    \includegraphics[width=0.7\columnwidth]{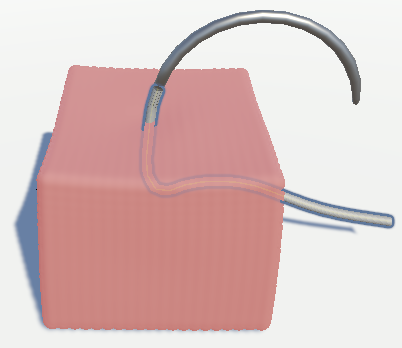}
    \caption{Simulated suture in soft tissue.}
    \label{fig:suture_in_tissue}
\end{figure}

While a large number of studies have explored the simulation of surgical sutures and biomedical soft tissue independently, only a few have considered the complete soft-tissue suturing scenario, including the contact between sutures and deformable tissue during suture insertion.
Some frameworks primarily focus on rigid-body capabilities, and the suturing scenes do not consider soft-tissue \cite{munawar2022OpenSimulationa,wu2024SurgicAIHierarchical}.
Simulators that consider more realistic scenarios often need to include different maths or solvers for both the suture and the tissue.
However, modelling contact during suture insertion within tissue, as well as the coupling between two solvers, presents additional challenges, even when compared to regular rope-soft-body contact.
In \cite{maciel2011SurgicalModelviewcontroller}, the tissue is simulated with FEM, while the suture thread uses a 1D mass-spring model kinematically constrained to the needle path.
This ignores the dynamic interaction between the suture and the tissue, and treats the suture as a ``light'' object that deforms entirely according to the needle path through the tissue.
In \cite{guebert2009SuturingSimulation}, sutures are modelled as serially-linked beam elements using FEM and interact with hexahedral FEM tissue.
Although both use FEM, insertion contact still requires separate modeling: bilateral constraints keep the needle shaft and suture thread within the path created by the needle tip.

A simpler approach that requires less contact modelling and coupling between different simulation solvers is to use PBD for both tissue and suture simulation \cite{yu2020RealtimeSuturing,westergaard2026RealtimeHapticbased}.
Suture-in-tissue can be modelled with constraints that directly fall under the regular PBD framework.
For example, authors of \cite{yu2020RealtimeSuturing} use PBD for modelling both tissue and suture and constrain suture particles so that they may slide along the needle path but cannot move away in orthogonal directions during suturing insertion.
However, although PBD sutures are often acceptable, tissue realism in PBD relies on constraints and non-physical parameters rather than true physical modeling.
This can introduce additional challenges for surgical robotics research, such as PBD parameter tuning to match the simulation with real-world tissue behavior.

To better simulate surgical suturing and allow robot learning with parallel simulations, this paper builds on previous work \cite{ou2025CRESSimMPM} and introduces a novel suturing simulation environment using sutures modelled by PBD and soft bodies modelled by the material point method (MPM) while considering dynamic contact, including frictional and drag forces.
The usage of MPM soft tissue preserves physical realism compared to using PBD tissue.
The main contributions of this work are:
\begin{itemize}
    \item We extend our previous CRESSim-MPM framework with a PBD-based suture model to support flexible suture deformation in surgical suturing, as in Fig.~\ref{fig:suture_in_tissue}.
    \item We introduce a novel contact coupling method between the PBD suture and the MPM soft tissue, enabling stable and realistic suture-tissue interactions.
    \item We optimize the simulator for GPU execution, build a parallel RL environment, and validate it by training agents for autonomous suturing sub-tasks.
\end{itemize}
It is worth noting that the RL task focuses only on validating that the simulator enables training vision-based policies while simulating soft tissue and suture contact, without considering the complete suturing task.
This is important for developing agents that account for suture-tissue contact and for transferring to the real world in future work.

\section{RELATED WORK}

\subsection{Surgical Robot Simulation}

This work is related to the recent trend in developing surgical simulators and autonomous surgical robot agents using simulators.
To build realistic surgical simulators, earlier studies include AMBF \cite{munawar2019RealTimeDynamic,munawar2022OpenSimulationa}, ATAR \cite{enayati2018RoboticAssistanceasNeeded}, and V-Rep Simulator for the dVRK \cite{fontanelli2018VREPSimulator}.
More recent works, such as dVRL \cite{richter2020OpenSourcedReinforcement}, UnityFlexML \cite{tagliabue2020SoftTissue}, AMBF-RL \cite{varier2022AMBFRLRealtime}, LapGym \cite{scheikl2023LapGymOpen}, ORBIT-Surgical \cite{yu2024OrbitSurgicalOpenSimulationa}, Surgical Gym \cite{schmidgall2024SurgicalGym}, FF-SRL \cite{dallalba2024FFSRLHigh}, SurRoL \cite{xu2021SurRoLOpensource} introduced platforms with specific focus on surgical robot learning.
However, most of them consider rigid body dynamics or cannot simulate soft-tissue suturing easily and efficiently due to the use of FEM, as discussed in Section~\ref{sec:introduction}.
This work proposes an alternative approach by combining MPM soft tissue and PBD suture simulation, addressing some of the limitations in existing studies.

\subsection{Autonomous Suturing}
\label{sub_sec:aut_suturing}
Suturing typically consists of multiple interdependent sub-steps, including needle localization and grasping, suture thread detection, needle insertion, needle driving, needle extraction, and knot tying. However, much of the existing research has focused on automating only two or three of these components. 


Similarly, in this work we assume that the needle has already been grasped, and focus only on the needle insertion, driving, and extraction phases of suturing. Early approaches in this direction learn and replay open-loop trajectories from human demonstrations \cite{schulman2013case, zhou2024suturing}, which lack feedback during execution and are sensitive to deviations from the demonstrated motion.
Subsequent methods incorporate real-time geometric planning \cite{ozguner2021visually} or IL \cite{huang2024visuomotor} to improve robustness, yet still rely on simplified planar or rigid tissue models that neglect internal physics of soft tissue, such as deformation forces and resistance.
More recent frameworks employ RL to address longer-horizon suturing tasks and needle–suture manipulation \cite{wu2024SurgicAIHierarchical, yu2024OrbitSurgicalOpenSimulationa}. Although these approaches improve realism through physics-based simulation and task decomposition, they do not explicitly model internal force-based needle–tissue interactions. Moreover, \cite{wu2024SurgicAIHierarchical} was not able to perform needle insertion using both PPO and SAC. In contrast, \cite{barnoy2021applying} incorporates direct needle–tissue interaction modeling via hybrid batch reinforcement learning, but is constrained by limited parallelism and CPU-based training in the da Vinci Skills Simulator. Finally, \cite{haworth2025suturebot} demonstrates a compelete suturing using various IL approaches, but training was conducted in real-world settings with limited environmental variability.



\section{ROPE AND TISSUE SIMULATION}

\subsection{Position-Based Dynamics}

In position-based dynamics (PBD), a continuum body is discretized into particles.
Consider each particle $i$ with its mass $m_i$, position $\mathbf{x}_i$ and velocity $\mathbf{v}_i$. The equation of motion is derived from Newton's law $\dot{\mathbf{v}}_i = \mathbf{f}_i / m_i,$
where $\mathbf{f}_i$ is the total force on the particle.
We can apply symplectic Euler integration to solve the system at each step:
\begin{equation}
	\label{eqn:app_pbd_integration}
    \begin{split}
        \mathbf{v}_i(t + \Delta t) = \mathbf{v}_i(t) + {\Delta t} \frac{1}{m_i} \mathbf{f}_i(t), \\
        \mathbf{x}_i(t + \Delta t) = \mathbf{x}_i(t) + {\Delta t} \mathbf{v}_i(t + \Delta t)
    \end{split}
\end{equation}

Instead of relying on the physical relationship between deformation and internal forces, kinematic constraints are used in PBD.
For instance, two neighboring particles can form a spring-like constraint.
PBD constraints are solved by projecting the predicted particle positions from (\ref{eqn:app_pbd_integration}) on the constraint manifold to correct the prediction.
The predicted particle positions are iteratively corrected by moving them toward satisfying all the constraints.
Consider a total of $M$ constraints:
\begin{equation}
	C_1(\mathbf{x}) \succ 0, \quad C_2(\mathbf{x}) \succ 0, \quad  \dots \quad
	C_M(\mathbf{x}) \succ 0,
\end{equation}
where $\mathbf{x}$ is the concentrated vector of all the particle positions.
We can employ the non-linear Gauss--Seidel approach to solve the positional corrections.
Specifically, we want to find a correction $\Delta \mathbf{x}$ such that $C(\mathbf{x} + \Delta \mathbf{x}) \succ 0$.
The constraints are linearized individually, approximated by
\begin{equation}
	\label{eqn:app_pbd_constraint}
	C(\mathbf{x} + \Delta \mathbf{x}) \approx C(\mathbf{x}) + \nabla C(\mathbf{x}) \Delta \mathbf{x} \succ 0.
\end{equation}
If $\Delta \mathbf{x}$ is restricted to be in the direction of $\nabla C$, $\mathbf{x}$ will move toward converging to satisfying the constraints:
\begin{equation}
	\label{eqn:app_pbd_x_correction}
	\Delta \mathbf{x} = - \lambda \mathbf{M}^{-1} \nabla C(\mathbf{x}),
\end{equation}
where $\mathbf{M} = \operatorname{diag}(m_1, m_2, \dots, m_N)$ for all $N$ particles.
Substituting (\ref{eqn:app_pbd_x_correction}) into the equality version of (\ref{eqn:app_pbd_constraint}) yields $\lambda$:
\begin{equation}
    \label{eqn:pbd_lambda}
    \lambda = \frac{C(\mathbf{x})}{\nabla C(\mathbf{x})^{\mathrel{T}} \mathbf{M}^{-1} \nabla C(\mathbf{x})}.
\end{equation}
This works for the equality constraints directly. Inequalities are handled by checking whether $C(\mathbf{x}) \ge 0$ to simply skip it in the iteration.

The extended PBD (XPBD) method additionally adds compliance to the constraints, achieving stiffness that is independent of the simulation time step, unlike in PBD.
Each constraint is augmented with a compliance parameter $\alpha \ge 0$. The new Lagrange multiplier increment can be computed as
\begin{equation}
    \Delta \lambda = \frac{-C(\mathbf{x}) - \alpha \lambda^{(k)}}{\nabla C(\mathbf{x})^{\mathrel{T}} \mathbf{M}^{-1} \nabla C(\mathbf{x}) + \alpha}.
\end{equation}
Here, $\lambda^{(k)}$ denotes the Lagrange multiplier value at the $k$-th iteration. The positional correction is then given by
\begin{equation}
    \Delta \mathbf{x} = \mathbf{M}^{-1} \nabla C(\mathbf{x}) \Delta \lambda ,
\end{equation}
and the multiplier is updated as $\lambda^{(k+1)} = \lambda^{(k)} + \Delta \lambda.$
Setting the compliance parameter $\alpha = 0$ results in XPBD being equivalent to the PBD formulation in (\ref{eqn:app_pbd_x_correction}) and (\ref{eqn:pbd_lambda}).

\subsection{PBD Suture Simulation}

A surgical suture is a rope-like object that can be discretized into connected particles and simulated using PBD.
Ignoring particle orientation, we use only distance constraints between neighboring particles and bending constraints among triplets of adjacent particles.

Mathematically, for neighbouring particles $i$ and $i+1$,
\begin{equation}
    C(\mathbf{x}_{i}, \mathbf{x}_{i+1}) = \lVert \mathbf{x}_{i} - \mathbf{x}_{i+1} \rVert - d_0 = 0,
\end{equation}
where $d_0$ is the constrained distance between the particles.
For triplets of adjacent particles $i-1$, $i$ and $i+1$, bending constraints are formed as
\begin{equation}
    C(\mathbf{x}_{i-1}, \mathbf{x}_{i}, \mathbf{x}_{i+1}) = \theta(\mathbf{x}_{i-1}, \mathbf{x}_{i}, \mathbf{x}_{i+1}) - \theta_0 = 0,
\end{equation}
where
\begin{equation}
    \theta(\mathbf{x}_{i-1}, \mathbf{x}_{i}, \mathbf{x}_{i+1}) = \arccos\left( \frac{\mathbf{x}_{i-1} - \mathbf{x}_{i}}{\lVert \mathbf{x}_{i-1} - \mathbf{x}_{i} \rVert} , \frac{\mathbf{x}_{i+1} - \mathbf{x}_{i}}{\lVert \mathbf{x}_{i+1} - \mathbf{x}_{i} \rVert} \right),
\end{equation}
and $\theta_0$ is the rest angle.
Both types of constraints are formed as XPBD constraints with compliance parameters.

This approach allows simulating rope-like objects without the need to consider particle orientations.
Alternative approaches that lead to more accurate simulations account for thread twisting constraints and add higher computational complexity.
Furthermore, the current implementation does not consider self-collision, which can be added with additional constraints.
Implementing these constraints and integrating them into our framework is straightforward if needed in future work, such as for simulating knot-tying.

\begin{figure}[t]
    \centering
    \includegraphics[width=0.5\columnwidth]{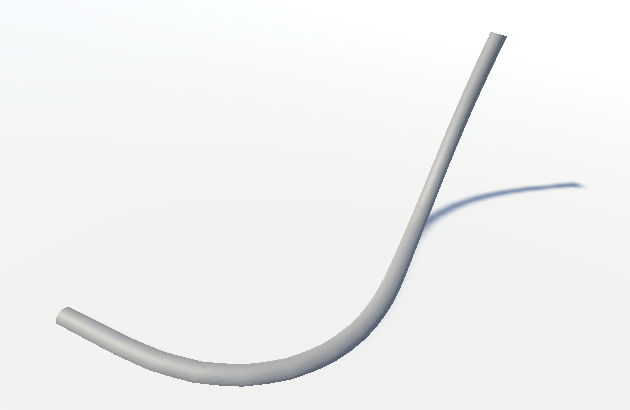}
    \caption{PBD suture simulation aligned with a rigged cylinder for visual representation.}
    \label{fig:pbd_suture_demo}
\end{figure}

Visually, a narrow cylinder rigged with bones is used to represent and animate the suture.
At each time step, the bones are aligned with the PBD particle positions to produce a visual simulation of the suture.
An example of the simulated suture is shown in Fig.~\ref{fig:pbd_suture_demo}.

\subsection{MPM Soft Tissue and Needle Contact Coupling}

Existing work, such as \cite{munawar2022OpenSimulationa}, has explored the use of PBD for suture simulation in surgical scenes. However, most of the existing studies are limited to contact models between PBD particles and rigid body solvers, and assume suturing through rigid apertures.
This work aims to achieve more realistic surgical simulation by enabling interaction between sutures and deformable soft tissue modeled using MPM.
This requires a novel contact modelling approach between the PBD suture and the MPM soft body.

In the MPM, a soft body is discretized into Lagrangian material points (moving particles) to store local material data, such as mass, velocity, stress tensor, and deformation gradient, while the deformation computation happens in an Eulerian (fixed) grid.
The MPM formulation follows 4 steps: \textbf{(1) Particle to grid (P2G)}: Material data at the particles are distributed to the grid using an interpolation shape function. \textbf{(2) Grid update}: the momentum of each grid node is integrated. \textbf{(3) Grid to particle (G2P)}: The updated node momenta are interpolated back to the particles. Material properties such as the deformation gradient and stress are also updated. \textbf{(4) Reset grid}: reset grid node data to zeros. 

\begin{figure}[ht]
    \centering
    \includegraphics[width=1\columnwidth]{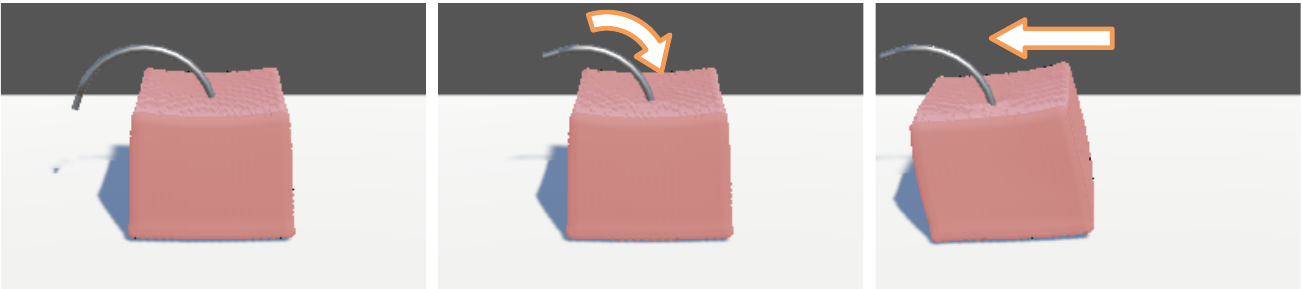}
    \caption{Arc-shaped suture needle in contact with MPM soft body.}
    \label{fig:arc_needle_in_tissue}
\end{figure}

Our previous work \cite{ou2025CRESSimMPM} introduced CRESSim-MPM, a CUDA-accelerated MPM simulation library designed for surgical soft-tissue simulation.
We also proposed a novel coupling strategy between a rigid needle and an MPM-based deformable body.
The needle is modeled as a 1D manifold (\eg an arc) when in contact with the soft tissue, allowing insertion and dragging, as indicated by the downward and leftward arrows in Fig.~\ref{fig:arc_needle_in_tissue}, respectively. Across all experiments, the tissue model was parameterized using Lamé elastic constants $\lambda = 714285.8$ and $\mu = 178571.4$, where $\lambda$ represents the volumetric stiffness and $\mu$ represents the shear stiffness.

\subsection{Coupling between MPM Soft Tissue and PBD Suture}

\begin{figure}[ht]
    \centering
    \includegraphics[width=0.8\columnwidth]{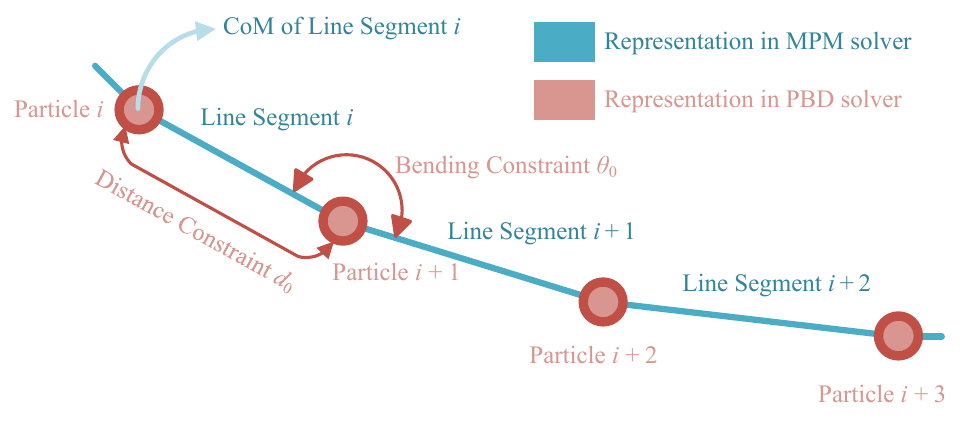}
    \caption{Suture representations in both PBD and MPM solvers for contact coupling between PBD suture and MPM soft body.}
    \label{fig:suture_mpm_contact}
\end{figure}

We can formulate the contact between the PBD suture and the soft tissue in the following manner: we construct virtual line segments between neighboring PBD particles as 1D geometries, similar to the needle representation, and constrain their motion accordingly, as shown in Fig.~\ref{fig:suture_mpm_contact}.

Specifically, for particles $i$ and $i+1$, we create a \texttt{ConnectedLineSegments} geometry in CRESSim-MPM to apply contact forces from the PBD suture to the MPM soft tissue.
To achieve two-way coupling, momentum changes induced by contact are applied as impulses from the MPM tissue back to the virtual line segments.

As these line segments are virtual from the perspective of the PBD solver, we found it sufficient to apply the resulting impulse directly to particle $i$ in most cases.
This corresponds to assuming that the center of mass (CoM) of the virtual line segment coincides with particle $i$, as shown in Fig.~\ref{fig:suture_mpm_contact}.

\begin{figure}[ht]
    \centering
    \includegraphics[width=1\columnwidth]{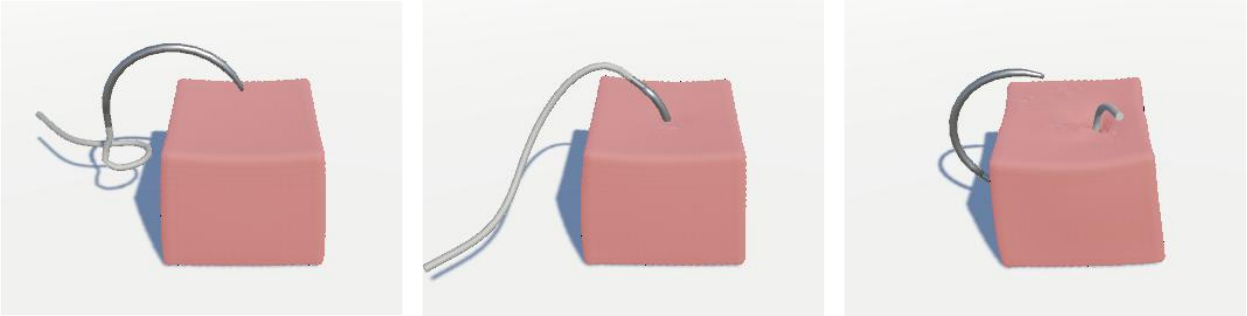}%
    \caption{Snapshots from a suturing demo scene}
    \label{fig:suturing_demo}
\end{figure}

On the MPM solver side, contact resolution is performed during the Eulerian grid update.
The resulting impulses are recorded and applied to the PBD particles after the MPM solver step and before the next PBD solver step.
Example snapshots of suturing in soft tissue are shown in Fig.~\ref{fig:suturing_demo}.

\subsection{Parallel Simulation}

The capability to simulate many instances of similar scenes is important for robot learning.
PBD-based thread simulation is not computationally expensive, even without GPU acceleration, especially due to the relatively small number of particles used to model rope-like objects. As a result, a large number of threads can be simulated in parallel, making the MPM soft tissue simulation the primary bottleneck in large-scale parallel simulation.

To better utilize GPU resources, we further introduce improvements to CRESSim-MPM. In particular, \texttt{Scene}-related data transfers and solver computations are separated into multiple CUDA streams, leading to improved GPU utilization across solver stages by overlapping the stepping of different \texttt{Scene}s under certain settings.
In practice, distributing \texttt{Scene}s to two CUDA streams on an NVIDIA RTX 4090 usually results in improved performance compared to using only one stream.
More detailed simulation performance evaluations will be presented in Section~\ref{subsec:simulation_performance}.

\begin{figure}[h]
    \centering
    \includegraphics[width=1\columnwidth]{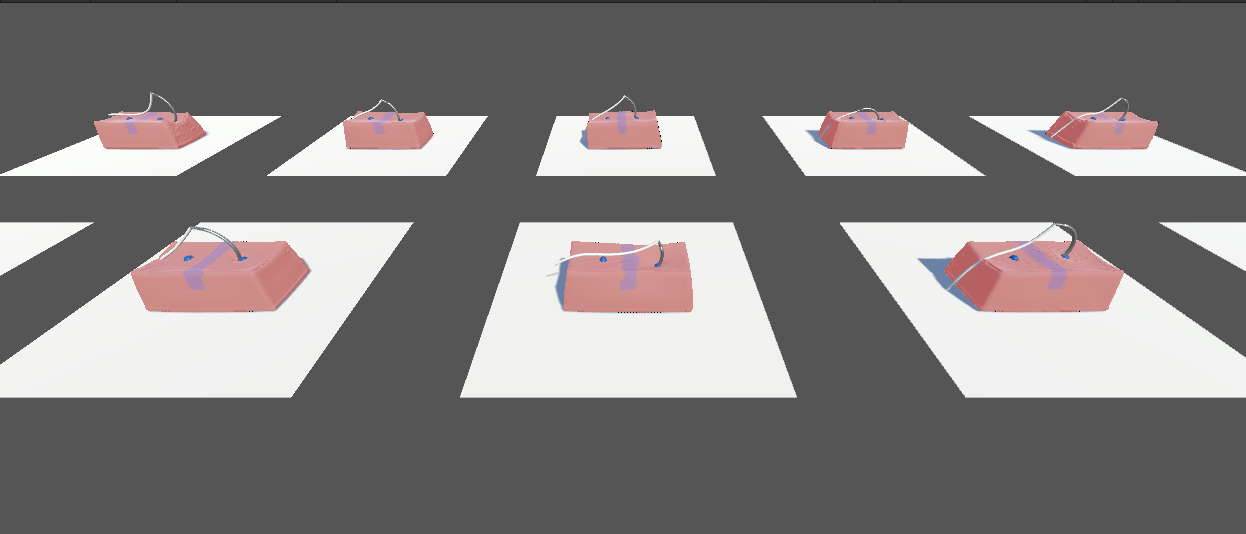}
    \caption{Multiple training parallel environments.}
    \label{fig:multiple_training}
\end{figure}

\section{REINFORCEMENT LEARNING ENVIRONMENT FOR SURGICAL SUTURING}

\subsection{Task Definition}
\label{sec:task_definition}

In this work, we develop a simplified surgical suturing RL task to demonstrate that autonomous agents can be trained using the simulation framework.

The objective is for a robotic manipulator to autonomously execute key sub-tasks of a suturing motion, including needle insertion at an entry marker, needle navigation inside the tissue, and needle extraction at a specified exit marker. In this study, we do not use a surgical robotic manipulator agents to grasp or manipulate the needle, but instead, the needle itself is treated as the agent. Therefore, we make several simplifying assumptions: the needle is already grasped prior to task execution, the suturing motion is confined to a planar workspace, and no re-grasping or hand-off of the needle is performed. The suture thread is attached to the needle end and therefore follows the needle path in the tissue.

The reward function is divided into three sequential phases, each corresponding to a separate sub-task of the suturing motion.
Let scalars \(w_e, w_\theta, w_x, w_T\) and \(b_e, b_x, b_n > 0\) be the manually specified weights and reward scales for the three phases.
In Phase~1, we guide the needle tip to the entry marker using a distance-based shaping term and a success bonus. Let \(t\) denote the time step. We define \(d_{\text{entry}}(t)\) as the Euclidean distance between the needle tip and the entry marker. In addition, we introduce an indicator function \(\mathbb{I}_{e(t)} \in \{0,1\}\), which equals to 1 when the needle tip reaches the entry marker at time \(t\), and 0 otherwise. This reward is
\begin{equation}
r_t^{(1)} 
= -\,w_{e}\, d_{\text{entry}}(t)
  + b_{e}\,\mathbb{I}_{e(t)} .
\end{equation}

\begin{figure}[ht]
    \centering
    \includegraphics[width=1.0\columnwidth]{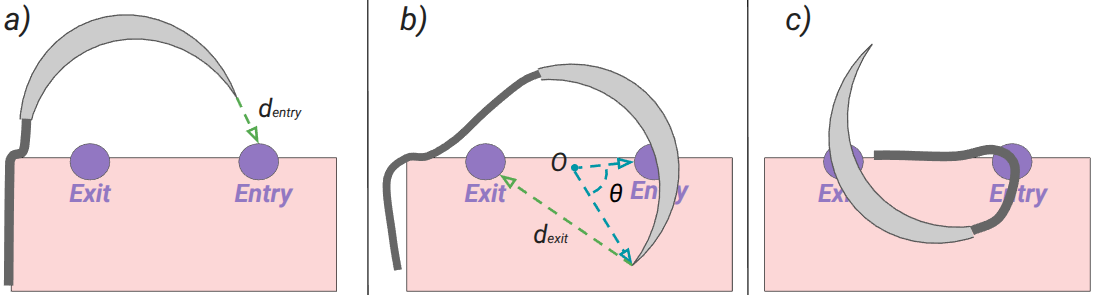}
    \caption{Phases of the suturing sub-task in side view: (a) needle insertion, (b) needle driving, (c) needle extraction.}
    \label{fig:suture_angle}
\end{figure}

Once the needle is in the tissue, Phase~2 encourages rotational needle driving while moving towards the exit marker.
Let \(o\) denote the needle’s arc intersection point (needle origin), and let \(e\) denote the entry marker.
We define the angle \(\theta(t)\) between vectors \(\overrightarrow{oe}\) and \(\overrightarrow{op_{\text{tip}}(t)}\), where \(p_{\text{tip}}(t)\) is the needle-tip position (Fig.~\ref{fig:suture_angle}). 
Minimizing \(\theta(t)\) encourages the needle to rotate about its origin and follow the desired arc through the tissue. Additionally, we panalize any translational motion inside the tissue to reduce needle sliding with \(v_{\text{trans}}(t)\). The Phase~2 reward is:

\begin{equation}
r_t^{(2)} 
= \,w_{\theta}\,\theta(t)
  + w_{x}\, d_{\text{exit}}(t)
  + b_{x}\,\mathbb{I}_{\text{x}(t)}
  - w_{T}\,\big\lVert v_{\text{trans}}(t)\big\rVert.
\end{equation}

Here, \(d_{\text{exit}}(t)\) is the inverse Euclidean distance between the needle tip and the exit marker, while \(\mathbb{I}_{x(t)} \in \{0,1\}\) is an indicator for when the needle tip reaches the exit marker. 

Finally, Phase~3 focuses on accurately extracting the needle from the exit marker. In this phase, we provide a terminal reward only upon successful task completion.
With an indicator function \(\mathbb{I}_{\text{f}(t)} \in \{0,1\}\):
\begin{equation}
r_t^{(3)} 
= b_{n}\,\mathbb{I}_{\text{f}(t)}.
\end{equation}

At each time step, the environment returns the reward corresponding to the active phase, \ie \(r_t = r_t^{\{1,2,3\}}\).

\subsection{Training Settings}
We utilize Unity’s ML-Agents Toolkit, which enables training with both proximal policy optimization (PPO) and Soft Actor-Critic (SAC). The simulation runs with a fixed time step of 0.004 s, while the decision requester is configured to issue actions every 20 simulation steps. At each decision step, the agent observes a $250 \times 250$ RGB image, along with the needle’s position and orientation. Observations are stacked over the current and two previous time steps and concatenated into a single vector. The inclusion of both visual and proprioceptive observations is important, since once the needle is inserted into the tissue, visual information alone is insufficient to infer its interaction with the tissue.

At the start of each episode, the tissue is initialized on a planar surface with a size of $2 \times 1\times 1$ ($W \times H \times D$) expressed in \textbf{Unity units (UU)}. In our simulation, 1 UU is interpreted as 1 meter; however, this unit-to-meter mapping can be rescaled when matching the simulation to real-world measurements.
The needle is spawned above it (Fig.~\ref{fig:suture_succ_example}a) and its initial position and orientation are randomized within bounded limits. To detect collisions between the needle tip and marker points, we used Unity’s collision detection system, which continuously checks whether a collision is triggered by the tip. Each episode is limited to a maximum of 4,000 environment steps.
We use a learning rate of 0.0003, a batch size of 1024, a replay buffer size of 10240, a constant entropy regularization coefficient of 0.01, and a discount factor of 0.95. To accelerate training, we run 10 parallel simulation environments concurrently. Both training and inference are performed on an NVIDIA RTX 4090 GPU.

\section{RESULTS}

\subsection{Simulation Performance}
\label{subsec:simulation_performance}
To quantitatively evaluate the simulation performance, an example scene is built in Unity for profiling.
This scene contains 10 parallel areas, each including a PBD suture with 20 particles and an MPM soft body with 20,000 material points.
The 10 parallel areas correspond to 10 \texttt{Scene}s in the CRESSim-MPM library.
The simulation step size is 0.004 seconds and the MPM integration step is 0.002 seconds, there are exactly two integration steps in the MPM solver for each \texttt{FixedUpdate} in Unity.

\textbf{Usage of CUDA streams: }
We show that using separate CUDA streams in CRESSim-MPM increases the computational efficiency of MPM simulation.
The example scene is run with a different number of CUDA streams used, and the MPM physics computation time is recorded.
For better results, trials are conducted without rendering, which is also computationally heavy and competes with the MPM solver for GPU resources.
The results over a 20-step window for each trial are shown in Table~\ref{tab:cuda_streams_mpm_performance}.
There is a significant performance improvement when increasing the number of CUDA streams from one to two.
However, using more than two streams yields limited additional gains, likely due to bottlenecks in other GPU resources, such as register pressure.

\renewcommand{\arraystretch}{1.3}
\begin{table}[ht]
\caption{MPM Solver Step Times When Using Different Numbers of CUDA Streams.}
\label{tab:cuda_streams_mpm_performance}
\centering
\begin{tabular}{cc}
\hline
\multicolumn{1}{l}{No. of CUDA streams} & \multicolumn{1}{l}{MPM physics step times (mean $\pm$ std, ms)} \\ \hline
1                                          & $5.85 \pm 0.15$                                                \\
2                                          & $2.10 \pm 0.13$                                                \\
3                                          & $1.97 \pm 0.17$                                                \\
4                                          & $1.70 \pm 0.20$                                                \\ \hline
\end{tabular}
\end{table}

\textbf{Computation time of individual components: }
We are also interested in the computational cost of each major component within a simulation step under standard simulation settings with soft-body rendering. To evaluate this, we collect computation times for MPM physics, PBD rope physics and visual alignment, and MPM soft-body rendering data computation at each step over a 20-step window.
Two CUDA streams are used.
The results are shown in Table~\ref{tab:individual_times}.
It is worth noting that the rendering of the MPM soft body treats material points as individual points with color information and follows the screen-space rendering technique described in \cite{ou2025learning}.
This is also computationally heavy and can compete with the MPM solver for GPU resources, which may explain why MPM physics computation is significantly slower than that reported in Table~\ref{tab:cuda_streams_mpm_performance}, although two CUDA streams are used in both cases.
PBD-related computation is fast despite its CPU implementation and is not a bottleneck in the overall simulation.

\begin{table}[ht]
\caption{Computation Times of Major Components.}
\label{tab:individual_times}
\centering
\begin{tabular}{lc}
\hline
Component                        & \multicolumn{1}{l}{Computation times (mean $\pm$ std, ms)} \\ \hline
MPM physics                      & $4.96 \pm 0.35$                                            \\
PBD physics and visual           & $0.63 \pm 0.07$                                            \\
MPM rendering                    & $0.60 \pm 0.70$                                            \\ \hline
\end{tabular}
\end{table}


\subsection{Training Results}

\begin{figure}[h]
    \centering
    \includegraphics[width=0.8\columnwidth]{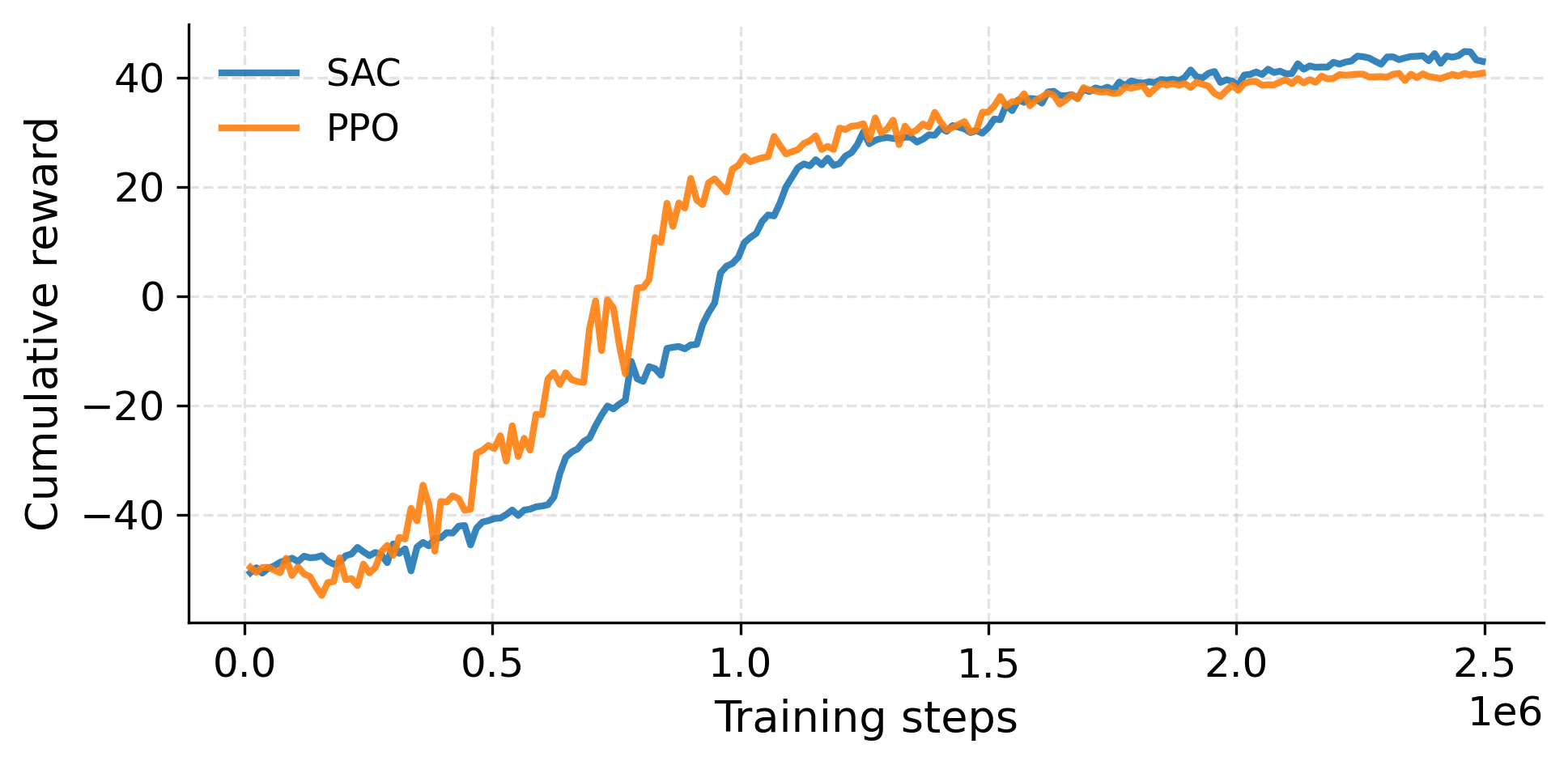}
    \caption{Learning curves using PPO and SAC trained on the suturing task.}
    \label{fig:training_curve}
\end{figure}

Fig.~\ref{fig:training_curve} shows the learning curve for agents trained with both SAC and PPO. These results indicate consistent policy improvement and stable convergence during training.

\begin{figure}[ht]
    \centering
    \includegraphics[width=0.8\columnwidth]{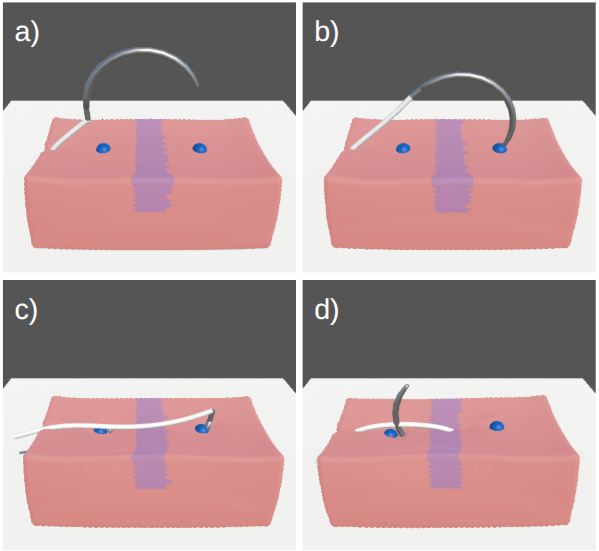}
    \caption{Result of the trained RL policy performing suturing:
    (a) random initial needle position,
    (b) needle tip penetrating the tissue at the entry point,
    (c) needle tip reaching the exit point,
    and (d) completing suturing.}
    \label{fig:suture_succ_example}
\end{figure}

We further evaluate the trained agents in the simulated environment by measuring task success rates across two phases: (1) needle insertion, defined as the needle entering the designated entry point within a specified distance threshold, and (2) needle extraction, defined as the needle exiting the tissue through the designated exit point (Fig. \ref{fig:suture_succ_example}). Results from 100 different episodes are recorded and summarized in Table \ref{tab:suturing_thresholds}.

\begin{table}[ht]
\centering
\caption{Success rates for needle entry and exit under different distance thresholds.}
\label{tab:suturing_thresholds}
\begin{tabular}{lccc}
\hline
\textbf{Metric} & \textbf{0.075 UU} & \textbf{0.12 UU} & \textbf{0.18 UU} \\
\hline
Entry Point Success Rate (\%) & 80 & 85 & 91 \\
Exit Point Success Rate (\%)  & 68 & 74 & 85 \\
\hline
\end{tabular}
\end{table}

In Table~\ref{tab:suturing_thresholds}, the reported thresholds are expressed in UU. As mentioned earlier, Unity Units (UU) refer to the engine’s native, dimensionless units. We use the marker centers as ground-truth reference points, and upon needle tip extraction from a marker, we record the Euclidean distance between them. A distance of 0.075~UU corresponds to the diameter of the entry and exit markers; therefore, successful insertion or extraction within this threshold is considered the most accurate.
The larger thresholds (0.12~UU and 0.18~UU) are used to evaluate how close the needle tip approaches the markers, with 0.12~UU indicating near-target performance and 0.18~UU representing a looser but still acceptable region.
Across 100 episodes, the agent successfully inserts the needle within the most accurate region (0.075~UU) in 80 cases. As expected, the success rate increases with larger thresholds, reaching 91\% within the 0.18~UU region, indicating that in only 9 out of 100 episodes the needle deviates significantly from the entry point at the beginning of the task.
Exit performance is comparatively lower than entry performance. This behavior is primarily attributed to the limited visual feedback of the needle once it is inside the tissue. Therefore, after insertion, the needle mainly follows the trajectory determined prior to entering the tissue. Despite this limitation, the agent achieves highly accurate exit performance (within 0.075~UU) in 68\% of the episodes and remains within the acceptable region of 0.18~UU in 85\% of the cases.

\section{LIMITATIONS AND FUTURE WORK}

The RL environment is built mainly for validating the proposed simulation framework and demonstrate that autonomous agents can be trained using it.
As mentioned in Section \ref{sec:task_definition}, we focus only on the needle insertion, driving, and extraction phases of suturing, while simplifying the action space and constraining the needle motion to a single plane. However, practical robotic suturing would require explicit modeling of needle re-grasping prior to extraction, needle hand-off, and knot-tying, which is not supported in the current setup. Incorporating physical robotic manipulators capable of performing these sub-tasks is therefore necessary for real-world applicability. Future work will extend the simulation to a dual-arm robotic setup, such as the da Vinci Research Kit (dVRK) with two Patient Side Manipulators (PSMs), enabling coordinated control for learning complete suturing procedures. Additionally, self-collision for the PBD suture should also be implemented.

Additionally, as an alternative to PPO or SAC, IL could also be considered.
However, as discussed in Section~\ref{sub_sec:aut_suturing}, the performance of IL is bounded by the quality and coverage of the available demonstrations. Consequently, deviations from the demonstrated distribution can lead to significant performance drops, while RL presents greater adaptability. In our current framework, we introduced variability only in the initial needle position. In future work, however, we plan to further randomize parameters including tissue appearance and mechanical properties, needle geometry, and entry/exit marker locations, settings in which the advantages of RL are expected to become more prevalent. However, we acknowledge that extending the framework to full suturing would require algorithms capable of handling long-horizon tasks, including imitation learning–based approaches.




\section{CONCLUSIONS}

This work introduced a novel simulation environment for surgical suturing with dynamic suture-tissue contact, tailored for parallel robot learning with GPU acceleration.
Building on CRESSim-MPM, a PBD-based suture was integrated, and a new coupling strategy between PBD sutures and MPM soft tissue was developed to achieve plausible contact dynamics.
The simulator was optimized for GPU execution, and a preliminary parallel RL environment was created to demonstrate its suitability for robotic learning. Using this framework, autonomous suturing agents were successfully trained using RL. Overall, this work demonstrates the feasibility and promise of achieving autonomous surgical suturing through simulation-only training, providing a scalable foundation for future advances in surgical robot learning.

\bibliographystyle{IEEEtran}
\bibliography{Misc.bib,Suturing.bib,SurgicalSimulation.bib}

\addtolength{\textheight}{-12cm}   


\end{document}